\documentclass[runningheads]{llncs}
\usepackage[T1]{fontenc}
\usepackage{setspace}

\usepackage{algorithm}
\usepackage{algpseudocode}
\usepackage{amsmath}

\usepackage{amsfonts}
\usepackage{ragged2e}

\usepackage{graphicx}
\usepackage{blindtext}
\usepackage{multirow}

\usepackage{glossaries}

\makeglossaries

\newglossaryentry{latex}
{
    name=latex,
    description={Is a markup language specially suited
    for scientific documents}
}

\newacronym{ui}{UI}{User Interface}
\newacronym{mae}{MAE}{Mean Absolute Error}
\newacronym{srtime}{SRTime}{System Response Time}
\newacronym{bptime}{BPTime}{Back-end Processing Time}
\newacronym{rppg}{rPPG}{Remote Photoplethysmography}
\newacronym{bvp}{BVP}{Blood Volume Pulse}
\newacronym{hr}{HR}{Heart Rate}
\newacronym{spo2}{SpO\textsubscript{2}}{Oxygen Saturation}
\newacronym{hrv}{HRV}{Heart Rate Variability}
\newacronym{rr}{RR}{Respiration Rate}
\newacronym{bp}{BP}{Blood Pressure}
\newacronym{bpm}{bpm}{beats-per-minute}
\newacronym{sbp}{SBP}{Systolic Blood Pressure}
\newacronym{dbp}{DBP}{Diastolic Blood Pressure}
\newacronym{ppg}{PPG}{Photoplethysmography}
\newacronym{ptt}{PTT}{Pulse Transit Time}
\newacronym{roi}{RoI}{Region of Interest}
\newacronym{sdk}{SDK}{Software Development Kit}
\newacronym{he}{HE}{Histogram Equalization}
\newacronym{lime}{LIME}{low light image enhancement via illumination map estimation}
\newacronym{snr}{SNR}{Signal-to-Noise-Ratio}
\newacronym{bss}{BSS}{Blind Source Separation}
\newacronym{fir}{FIR}{Finite Impulse Response}
\newacronym{pca}{PCA}{Principal Component Analysis}
\newacronym{ica}{ICA}{Independent Component Analysis}
\newacronym{fft}{FFT}{Fast Fourier Transform}
\newacronym{dct}{DCT}{Discrete Cosine Transform}
\newacronym{svm}{SVM}{Support Vector Machines}
\newacronym{ac}{AC}{Alternating Current}
\newacronym{dc}{DC}{Direct Current}
\newacronym{ecg}{ECG}{electrocardiogram}
\newacronym{rmse}{RMSE}{Root Mean Square Error}
\newacronym{pat}{PAT}{Pulse Arrival Time}
\newacronym{pwv}{PWV}{Pulse Wave Velocity}
\newacronym{pcg}{PCG}{Phonocardiogram}
\newacronym{mlp}{MLP}{Multilayer Perceptron}
\newacronym{abp}{ABP}{Arterial Blood Pressure}
\newacronym{svr}{SVR}{Support Vector Regression}
\newacronym{toi}{TOI}{Transdermal Optical Imaging}
\newacronym{cnn}{CNN}{Convolutional Neural Network}
\newacronym{lstm}{LSTM}{Long Short Term Memory}
\newacronym{ssd}{SSD}{Single Shot multibox Detector}
\newacronym{hsv}{HSV}{Hue-Saturation-Value}
\newacronym{psd}{PSD}{Power Spectral Density}
\newacronym{ibi}{IBI}{Inter Beat Interval}
\newacronym{fps}{fps}{frames per second}
\newacronym{s}{secs}{seconds}
\newacronym{url}{URL}{Uniform Resource Locator}
\newacronym{bam}{BAM Lab}{Big Data Analytics and Management Laboratory}
\newacronym{iot}{IoT}{Internet of Things}
\newacronym{icu}{ICU}{Intensive Care Unit}
\newacronym{greb}{GREB}{Graduate Research Ethics Board} 

\begin{document}
\title{A Web Application for Experimenting and Validating Remote Measurement of Vital Signs}
\author{Amtul Haq Ayesha\inst{1}\orcidID{0000-0002-5895-7441} \and
Donghao Qiao\inst{1}\orcidID{0000-0003-1411-0705} \and
Farhana Zulkernine\inst{1}\orcidID{0000-0002-3326-0875}}
\authorrunning{A. Ayesha et al.}
\institute{Queen's University, Kingston Ontario, Canada \\
\email{20aha@queensu.ca, d.qiao@queensu.ca, farhana.zulkernine@queensu.ca}}
\maketitle              % typeset the header of the contribution

\begin{abstract}
With a surge in online medical advising remote monitoring of patient vitals is required. This can be facilitated with the \acrfull{rppg} techniques that compute vital signs from facial videos. It involves processing video frames to obtain skin pixels, extracting the cardiac data from it and applying signal processing filters to extract the \acrfull{bvp} signal. Different algorithms are applied to the \acrshort{bvp} signal to estimate the various vital signs. We implemented a web application framework to measure a person’s \acrfull{hr}, \acrfull{hrv}, \acrfull{spo2}, \acrfull{rr}, \acrfull{bp}, and stress from the face video. The \acrshort{rppg} technique is highly sensitive to illumination and motion variation. The web application guides the users to reduce the noise due to these variations and thereby yield a cleaner \acrshort{bvp} signal. The accuracy and robustness of the framework was validated with the help of volunteers.

\end{abstract}

\section{Introduction}
Measurement of vital signs like the \acrshort{hr}, \acrshort{hrv}, \acrshort{spo2}, \acrshort{bp}, stress and temperature are important to understand a patient's health status \cite{B49}. Presently, monitoring these vitals requires patients to either visit a clinical facility, or buy multiple devices such as the \acrshort{bp} monitor, oximeter, and thermometer which they must learn to use. Wearable sensor devices like smart watches are also available but patients must buy the reliable devices approved by Health Canada. Therefore, an alternative mode of remote vital signs monitoring with a single device (smartphone or web camera) will be beneficial as the users can measure their vitals at the comfort of being at home and without buying additional devices or receiving prior training on device usage. 

In the 1930s, Hertzman proposed the principle of \acrfull{ppg} \cite{B47}. In \acrshort{ppg} method, the skin is illuminated with light and in proportion to the volume of blood flowing through the tissues, a part of the light is absorbed by the tissues and the rest is reflected. From the reflected light, the \acrshort{bvp} signal is extracted, which is processed further to compute the \acrshort{hr} \cite{B1}. The first commercial oximeter based on \acrshort{ppg} was introduced in 1983 \cite{B48}. Oximeters contain a photodiode sensor which measures the intensity of reflected light. Based on this technique, many commercial devices are available today and are widely used to measure the HR and SpO2 \cite{B68}. Researchers have used \acrshort{ppg} signals obtained from contact \acrshort{ppg} sensor devices and analyzed them using machine learning algorithms to calculate \acrshort{hr} and \acrshort{bp} \cite{B16,B37}. With the popularity and wide use of camera based smartphones, researchers have used videos of fingertip and monitored changes in skin color over a period of time to extract the \acrshort{bvp} signal \cite{B29}. In recent years, the remote Photoplethysmography (\acrshort{rppg}) methods for measuring vital signs based on the principle of \acrshort{ppg} have gained momentum, which are referred to as \acrshort{rppg} methods \cite{B8}. These methods employ a contactless camera to capture face video for vital signs measurement under laboratory environment \cite{B19,B26} with controlled lighting conditions and no subject movements. These good quality videos without real life environmental noises result in clean \acrshort{bvp} signals and provide good measurement accuracy. However, many users are hesitant in using online systems to record their face videos. Therefore, the technology needs to be enhanced with privacy measures to work in real world use case scenario and validated using a large sample population having different physical traits and health conditions before it can be deployed in clinical care in Canada. 

\begin{figure}[!b]
\includegraphics[width=1.0\textwidth]{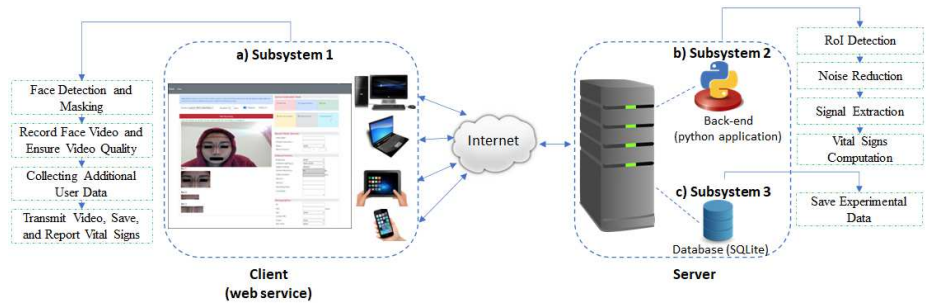}
\caption{Overview of the framework. It comprises of three subsystems: (a) front-end HTML application, (b) back-end processing python module and (c) SQLite database.} \label{fig:fig_framework}
\end{figure}

In this paper, (1) we present a web application framework with a back-end server as shown in Fig. \ref{fig:fig_framework} 
for remote web-based measurement of vitals signs namely \acrshort{hr}, \acrshort{hrv}, \acrshort{spo2}, \acrshort{rr}, \acrshort{bp} and stress in near real-time using a privacy preserving face video captured with a device camera. (2) We validate the \acrshort{rppg} technology using our web application in the real world environment with different sources of light, varying camera resolutions, multiple browsers, several devices, and networks. (3) Extensive research was done to explore existing \acrshort{rppg} methods \cite{B59} and improve the \acrshort{bvp} signal by diminishing motion and light noises encountered in real world environment and giving the user appropriate messages to capture a good quality video. In this version of the application, scalability and load balancing was not addressed. Instead, we focused mainly on validating the accuracy of the framework in the real world.

The rest of the paper is organized as follows. Background and related work are presented in Section \ref{sec:Background}. The web application framework is explained in Section \ref{sec:Framework} and its experimental validation is described in Section \ref{sec:Experiments}. Finally, Section \ref{sec:Conclusion} concludes the paper with an outline of future work directions.

\section{Background and Related Work} \label{sec:Background}
A web application is a software built using client-server architecture with the client side made accessible on the web to communicate with the users and obtain information that is transferred to the back end server for processing. The results can be reported back to the user. It can allow ubiquitous access to a wide range of users, and therefore, is ideal to validate the rPPG technology using a large number of volunteers. In this section we present some required concepts and literature review of the different methodologies we applied to implement a robust \acrshort{rppg} framework.

\subsection{\acrfull{rppg}} \label{section:rppg}
\acrshort{rppg} models estimate user vital signs from face videos using signal processing techniques and machine learning models \cite{B49}. The complete method consists of the following steps. 
\begin{enumerate}
    \item Detect \acrfull{roi}s: Identify face landmarks such as eyes, nose, lips, forehead, cheeks, and segment \acrshort{roi}s in the video frames to obtain the raw signal;
    \item Noise reduction: Improve the video quality to reduce the noise due to light and motion and thereby, improve the quality of the raw signal; 
    \item Signal extraction: From the \acrshort{roi}s of the improved video, \acrshort{bvp} signals are extracted based on change in pixel colors representing the periodicity of blood flow under the skin; 
    \item Vital signs computation: On the BVP signal, different computational pipelines are applied to calculate \acrshort{hr}, \acrshort{hrv}, \acrshort{spo2}, \acrshort{rr}, \acrshort{bp} and stress.
\end{enumerate} 

\subsection{Literature Review} \label{section:lit_review}
\paragraph{\textbf{RoI Detection:}}
The face is detected and suitable \acrshort{roi}s are segmented to extract a periodic \acrshort{bvp} signal. This signal is often dampened by motion artifacts owing to involuntary facial movement like blinking, twitching, smiling, and frowning \cite{B1,B26}. Therefore, it becomes necessary to choose a \acrshort{roi} that includes the least noise and the most cardiac information. Two methods are commonly used for extracting the \acrshort{roi}s on the face \cite{B1,B26}. The first method uses a face detector to segment the face with a bounding box. The second method predicts the coordinates of the facial landmarks, which can then aid in segmenting the \acrshort{roi}s.

\paragraph{\textbf{Noise Reduction:}}
Two major sources of noise that result in poor quality \acrshort{bvp} signal are, (a) inconsistent illumination and (b) movement \cite{B49}.

\textit{Illumination Noise:} In low light environment the skin cardiac data is not clearly visible, which affects the extracted PPG features \cite{B26}. Guo et al. \cite{B24} applied \acrfull{he} %and the \acrfull{lime} algorithm 
to the videos and found that the enhanced videos gave larger \acrshort{roi}s than the original ones and also improved the quality of the signal. Qiao et al \cite{B49} used \acrshort{he} to improve the lighting in the video when the background light was low. %Quellec et al. \cite{B27} improved the image quality by treating the image in the YCrCb color space. Leaving the Cr and Cb components unchanged, the Y component (known to represent luminance) is used to account for illumination variation. The image background is extracted by using a large size Gaussian kernel (standard deviation of 5 pixels) and is subtracted from the Y channel. The resultant image is converted to an RGB image. 

\textit{Motion Noise:} Rahman et al. \cite{B46} and Qiao et al. \cite{B59} used detrending filter and moving average filter to remove the stationary components and motion artifacts from the signal, respectively. Detrending helps attenuate the background intensity noise from the signal. Moving average filter computes the average of the datapoints between the video frames, thereby reducing the random noise yet retaining a sharp step response. The denoise filter helps in removing the jumps and steps in the signal caused by head movements such as rotation or shaking. %For motion elimination, Li et al. \cite{B26} and Haan et al. \cite{B25} divided the signal $g_{IR}$ into $m$ segments of equal length and computed the standard deviation of each segment. The top 5\% of the segments with the highest standard deviation are eliminated and the remaining segments are concatenated. Li et al. \cite{B26} applied the detrending filter to remove the low frequencies from the signal. To further eliminate the irrelevant frequencies, various bandpass filters were used. Li et al. \cite{B26} used \acrfull{fir} bandpass filters with cutoff frequency representing the normal \acrshort{hr}.

\paragraph{\textbf{Signal Extraction:}}
The individual face video frames are monitored over a period of time to track the changes in pixel color intensity to generate the \acrshort{bvp} signal. All the three-color channels namely red (R), green (G) and blue (B), contain pulsatile data. Wang et al. \cite{B80} utilized the data from the green channel while Poh et al. \cite{B50} used all the three-color channels. %To enhance the \acrshort{snr} of the raw signal, Burzo et al. \cite{B69} computed the spatial average of pixels along each color channel.

\paragraph{\textbf{Vital Signs Computation:}}
The different methods are briefly described below. %More elaborate information about the computation methods can be found in the related work as mentioned in the next subsection.

\textit{\acrshort{hr}:} The interval between the peaks of the time-domain \acrshort{bvp} signal indicates the \acrshort{hr} but this method is very sensitive to noise. The \acrshort{bvp} signal can be transformed into the frequency domain using Fast Fourier Transform \cite{B26}. The highest peak in the frequency spectrum is the fundamental frequency $f_{HR}$. Qiao et al \cite{B49} calculated \acrshort{hr} as $f_{HR}$ * 60.  %Deep learning models like DeepPhys \cite{B91} and HR-Net \cite{B92} extracted \acrshort{bvp} signals from a series of images in videos. %Shyam et al. \cite{B106} used \acrshort{ppg} signal from a wrist worn sensor and fed it to a deep neural network PPGnet, which consisted of \acrfull{cnn} and \acrfull{lstm} components. Brophy et al. \cite{B100} used a feed forward neural network to estimate \acrshort{hr} from \acrshort{ppg} signal obtained from fingertip sensor.

\textit{\acrshort{hrv}:} \acrfull{ibi} is the time period between the heartbeats. \acrshort{hrv} can be computed by calculating the time interval between two successive peaks in the \acrshort{bvp} signal \cite{B4}. Qiao et al. \cite{B49,B59} calculated $IBI=t_n-t_{(n-1)}$ where $t_n$ is the time of the $n\ th$ detected peak. They calculated \acrshort{hrv} according to Eq. \ref{eq:rmssd}, where $N$ is the number of \acrshort{ibi}s in the sequence.
\begin{equation}
HRV=\sqrt{\frac{1}{N-1}\sum_{i=1}^{N-1}(IBI_i-IBI_{i+1})^2}
\label{eq:rmssd}
\end{equation}

\textit{\acrshort{spo2}:} Based on the principle that the absorbance of Red (R) light and Infrared (IR) light by the pulsatile blood changes with the degree of oxygenation \cite{B93}, \acrshort{spo2} is calculated from the \acrshort{bvp} signal. The extracted \acrshort{bvp} signal obtained from the reflected light is divided into two parts: the \acrfull{ac} component resulting from the arterial blood and the \acrfull{dc} component resulting from the underlying tissues, venous blood, and constant part of arterial blood flow. The \acrshort{spo2} level in the blood can be calculated using Eq. \ref{eq: spo2}.

\begin{equation}
SpO_2=A-B\times\frac{AC_R/DC_R}{AC_{IR}/DC_{IR}}
\label{eq: spo2}
\end{equation}
where parameters $A$ and $B$ can be calibrated by using a pulse oximeter. Qiao et al \cite{B49} set 1 and 0.04 as the calibration parameters $A$ and $B$ respectively.

\textit{\acrshort{rr}:} Due to its non-stationary nature, estimating \acrshort{rr} from \acrshort{ppg} is challenging. Park et al. \cite{B102} extracted the dominant frequency from the \acrshort{bvp} signal, used an infinite impulse response filter to eliminate cardiac component, and then used adaptive lattice notch filter to estimate \acrshort{rr}. In this project, we estimated \acrshort{rr} by using a bandpass filter to retain frequencies in the range of 0.15 - 0.35 Hz, and the peak in the resultant signal times 60 was taken as the \acrshort{rr}.

\textit{\acrshort{bp}:} Non-linear regression models have shown good accuracy in estimating the \acrshort{bp} \cite{B36} proving that \acrshort{bp} and \acrshort{ppg} have a non-linear correlation. Shimazaki et al. \cite{B37} used autoencoders to extract the complex features that could be used as input to a four layer neural network. Viejo et al. \cite{B40} fed the amplitude and frequency of detected peaks to a regression model comprising a two layer feed forward network. Huang et al. \cite{B41} used the results from transfer learning on MIMIC II dataset with k-nearest neighbours for \acrshort{bp} prediction from face videos. Qiao et al. used a deep neural network with ResNet blocks and employed transfer learning by first training the network with finger \acrshort{ppg} data and then with face video \acrshort{rppg} data.

\textit{Stress:} \acrshort{hr} is an indicator of stress level. In this project, we calibrate the stress as follows: relaxed when \acrshort{hr} $<$ 67 \acrshort{bpm}, normal when 67-75 \acrshort{bpm}, low when 75-83 \acrshort{bpm}, medium when 84-91 \acrshort{bpm}, high when 92-100 \acrshort{bpm}, very high when 101-109 \acrshort{bpm} and extreme when \acrshort{hr} $>$ 109 \acrshort{bpm}.

An overview of the commercially available applications for estimating vital signs from face videos is illustrated in Table \ref{table:Apps_1}. These applications offer the solution to measure multiple vital signs with a single device which is why we implemented the same approach with improved noise reduction techniques.
\begin{table}[!b]
    \caption{Overview of existing commercial applications for remote measurement of vital signs.}
    \label{table:Apps_1}
        \centering
        \small
        \begin{tabular}{|l|l|c|c|c|c|c|c|c|c|c|c|}
        \hline
             \multirow{2}{*}{\textbf{App}}& \multirow{2}{*}{\textbf{Technology}} & \multirow{2}{*}{\textbf{Face}} & \multirow{2}{*}{\textbf{Finger}} &
             \multirow{2}{*}{\textbf{SDK}} & \multirow{2}{*}{\textbf{Free}} & \multicolumn{6}{c|}{\textbf{Vitals Measured}} \\ \cline{7-12}
                & & & & & & \textbf{\acrshort{hr}} & \textbf{\acrshort{hrv}} & \textbf{\acrshort{rr}} & \textbf{\acrshort{spo2}}& \textbf{\acrshort{bp}} & \textbf{Stress} \\
        \hline
            Anura \cite{B75} & \acrshort{toi} & \checkmark &  & & \checkmark & \checkmark &  &  & & \checkmark & \checkmark \\

        \hline
            Happitech \cite{B104} & \acrshort{ppg} & & \checkmark & \checkmark & \checkmark & & \checkmark & & & &   \\

        \hline
            Binah.ai \cite{B77} & \acrshort{ppg}, \acrshort{rppg} & \checkmark  & \checkmark & \checkmark &  & \checkmark & \checkmark & \checkmark & \checkmark & \checkmark & \checkmark\\

        \hline
            Veyetals \cite{B112} & \acrshort{ppg}, \acrshort{rppg} & \checkmark  & \checkmark & \checkmark & \checkmark & \checkmark  & \checkmark & \checkmark  & \checkmark & \checkmark & \checkmark  \\
        \hline

        \end{tabular}
\end{table}
\section{Web Application Framework} \label{sec:Framework}
The proposed web application framework version 1.0 is a python web framework having a client-server architecture which is composed of three subsystems as shown in Fig. \ref{fig:fig_framework}.
The server hosts the front-end application, manages resources, and delivers the back-end functionality including data processing, storage, and running the computational models for estimating and returning the vital signs namely \acrshort{hr}, \acrshort{hrv}, \acrshort{rr}, \acrshort{spo2}, \acrshort{bp}, and stress from face videos. We specifically focused on the computational models in this version, which will be demonstrated and evaluated through experiments. Load balancing and scalability will be addressed in future work. 

We used the Flask framework \footnote{\url{https://flask.palletsprojects.com/en/2.0.x/}}, which is a popular lightweight Python microframework with a built-in development server and support for unit testing. The framework also has a strong community support and excellent documentation which made it our choice for this application. %The processes implemented in the framework in each subsystem is illustrated in Fig. \ref{fig:fig_framework_flow}. 
%\begin{figure}[!b]
%   \centering
%   \includegraphics[scale=0.55]{Figures/Framework_Flowchart}
%    \caption{A flowchart of the processes implemented in the end-to-end framework.}
%    \label{fig:fig_framework_flow}
%\end{figure}

\begin{figure}%[!b]
   \centering
   \includegraphics[scale=0.8]{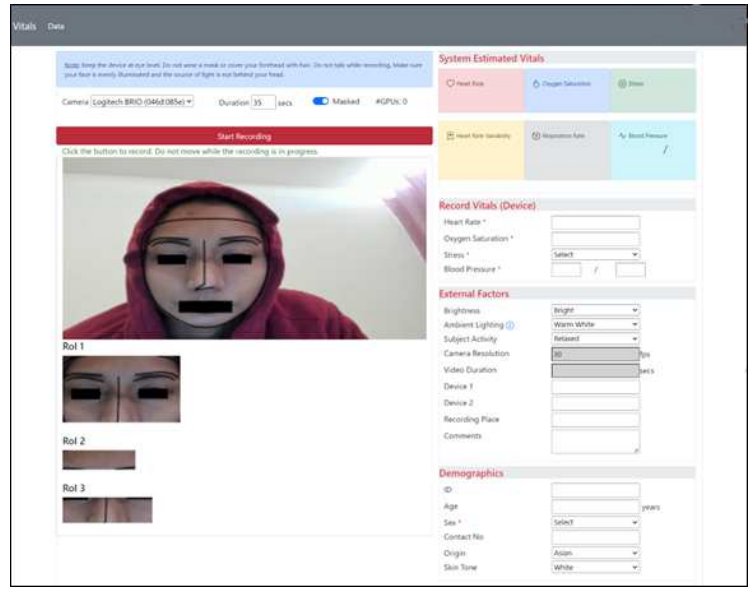}
    \caption{Front-end web interface}
    \label{fig:fig_UI}
\end{figure}

\subsection{Subsystem 1: Front-end Web Interface}
The front-end web interface shown in Fig. \ref{fig:fig_UI} can be accessed on the browser via a public web URL\footnote{\url{https://vital-signs-bamlab.tk/}}. It is built using HTML5, CSS, jQuery and Bootstrap. The front-end application requests camera access and upon receiving access, it starts capturing user's video. It anonymizes the user video by face masking, monitors the user position, sends the video to the back-end for computing vitals, displays the vitals on the user interface, and collects optional user data such as age, skin color, and other relevant information to enable further analysis as explained below.
\begin{enumerate}
    \item \textbf{Face Detection and Masking:} To protect user privacy, we apply face masking by detecting face landmarks such as eyes, lips, jawline, and nose, and covering them with another object called the face mask. Three APIs namely Haar-cascade classifier \cite{B22}, face-api.js \cite{B52}, and MediaPipe Face Mesh framework JavaScript API \cite{B65} were tested for face and facial landmark detection. We used the MediaPipe Face Mesh API as it made real-time predictions and gave superior user experience. We apply a simple mask by covering the mouth and eyes with black strips, and drawing black contour lines on the nose area, eyebrows, and face edges. A button on the user interface allows the user to turn face mask on or off. %For experimental analysis we recorded and evaluated both masked and unmasked videos. 
    \item \textbf{Record Face Video and Ensure Video Quality:} The device camera records the user's face video in MP4 format using the MediaRecorder\footnote{\url{https://developer.mozilla.org/en-US/docs/Web/API/MediaRecorder}} interface of the MediaStream Recording API. A bounding box around the face is computed to determine the face area. User distance from the camera is computed as the ratio of the face area to the video frame area. Tracking of the face landmarks throughout the video allows the system to detect too much movement (beyond 15 units of displacement of the bounding box) and stop the recording. If the user is too far from the camera or moving too much, the \acrfull{ui} displays a message to guide the user to reposition or stay steady to obtain a good quality video. 
    \item \textbf{Collect Additional User Data:} To validate the accuracy of the measurements for different parameters, some data is collected from the user using a form on the \acrshort{ui} as listed below.
        \begin{enumerate}
        \item Ground truth data measurements of \acrshort{hr}, \acrshort{spo2}, stress and \acrshort{bp} taken by the user or the researcher using Health Canada approved medical devices. The \acrshort{hrv} and \acrshort{rr} measurements are sensitive measurements which are typically measured at a clinical facility. In this study we did not record the \acrshort{hrv} and \acrshort{rr} ground truth values (future study) and used a benchmark dataset containing these values.
        \item Environmental data such as the brightness of the place (bright/dark), type of ambient lighting (warm white/ cool white/ daylight) causing illumination noise, and subject activity (relaxed/post exercise) causing motion noise, which affect the measurement accuracy. 
        \item User Profile such as name, age, sex, skin tone (white, yellow, brown, dark), and ethnicity (Asian, South-Asian, White Caucasian, African American, Hispanic) can be optionally provided by the user.
        \end{enumerate}
    \item \textbf{Transmit Video, Save, and Report Vital Signs:} Once the noise is acceptable, the recording is transmitted to Subsystem 2 for further processing and calculation of vital signs. Calculated measurements are displayed on the \acrshort{ui} in near real time. A \textit{Save} button allows all information to be saved on Subsystem 3, which can be accessed later for further studies.
\end{enumerate}

\subsection{Subsystem 2: Back-end Data Processing}
The back-end hosts a Python 3.8 application on the server. The videos received from Subsystem 1 are saved on the server, processed and analyzed to extract the raw \acrshort{ppg} signals and reduce noises due to changes in light intensity and motion to obtain a robust \acrshort{bvp} signal for higher accuracy of vital signs. We used the state-of-the-art methods from Qiao et al \cite{B49} for vital sign estimation as explained briefly in Section \ref{section:lit_review} and improve the noise filtering techniques (selecting the best RoI, guiding user by UI messaging) while validating the framework in real life environment.

\subsection{Subsystem 3: SQLite Database}
Due to the ease of installation, usage, and portability, we used the SQLite\footnote{\url{https://www.sqlite.org/index.html}} database version 3.12.2 as the repository on the back-end server to store the experimental data. It consists of a single relational table which saves time stamped data for each user session including the video file name, the vitals calculated by Subsystem 2, and the measured vitals entered by the user in Subsystem 1 along with the additional user and environmental details collected thruough the UI.

\subsection {Application Deployment}
To host the framework on a public server accessible by a URL, a Nginx web server and Gunicorn HTTP server is used, as the built-in Flask web server cannot be used for production. Nginx\footnote{\url{https://www.nginx.com/}} is a powerful open source web server that can handle reverse proxy, load balancing, security, scalability and HTTP caching. Gunicorn\footnote{\url{https://gunicorn.org/}} is a Python Web Server Gateway Interface (WSGI) HTTP server for UNIX.

%The deployment architecture is shown in Fig. \ref{fig:fig_architecture}. When the client requests the URL \url{https://vita-signs-bamlab.tk} a Domain Name System (DNS) lookup is performed to find the application server's IP address. Once the browser receives the IP address, it establishes a TCP connection with the server and sends a HTTP request with the recorded video. The Nginx engine acts as a reverse proxy server listening on port 80 of the server. Upon receiving an incoming client request, it passes the request to the web socket configured in the Gunicorn service. Gunicorn web socket binds to a  WSGI access point which is the entry point to the flask application. After receiving the request to calculate the vitals from the video, the flask application processes the video and computes the results. These results are sent back as response to Gunicorn which relays it to the Nginx which sends the HTTP response to the client.

\section{Experiments and Results} \label{sec:Experiments}
To demonstrate the usability and performance of the framework, participants were asked to use the application in real life environment. The vitals were measured simultaneously using medical devices Omron HEM-FL31 BP monitor and LOOKEE LK50D1A pulse oximeter.  We used the \acrfull{mae}, \acrfull{srtime}, and \acrfull{bptime} as the metric for evaluating the experimental results. The difference between the ground truth vital sign reading measured with a medical device and the vital sign value estimated by the framework, is the error. Average error value computed from multiple measurement sessions is called the Mean Absolute Error (\acrshort{mae}). The \acrshort{srtime} is the total time taken by the application to collect the video and report the results. The \acrshort{bptime} is the time taken by the back-end system to process the videos and report the results. We conducted experiments to validate the framework's: 
\begin{enumerate}
    \item \textbf{Accuracy:} Five volunteers in the age group of 13-40 years used the application at different times of the day and in different physical states on the Google Chrome browser of an Acer Aspire A315-55G laptop, with a Logitech UltraHD 4K webcam. The ground truth vital signs readings were in the range of 72-108 \acrshort{bpm}, 97-100\%, 94-114 mmHg, and 58-76 mmHg for \acrshort{hr}, \acrshort{spo2}, \acrshort{sbp}, and \acrshort{dbp} respectively. The results of this experiment are shown in Table \ref{table:accuracy}. %The vitals were measured simultaneously using medical devices Omron HEM-FL31 BP monitor and LOOKEE LK50D1A pulse oximeter, and the readings were in the range of 72-108 \acrshort{bpm}, 97-100\%, 94-114 mmHg, and 58-76 mmHg for \acrshort{hr}, \acrshort{spo2}, \acrshort{sbp}, and \acrshort{dbp} respectively. 
    \item \textbf{Robustness and Performance:} One volunteer used the application for 5 days with multiple sources of light and camera resolutions, and different internet networks, devices and browsers. The application was used three times in a day at 10:00, 13:00, and 16:00 hrs. The experimental results are shown in Table \ref{table:robustness} and \ref{table:performance}.
    \item \textbf{Workload Capacity:} The maximum number of users the application can support was tested using Google Chrome browser on different devices starting with 9 participants. 
\end{enumerate}

\textit{Discussion:} From Table \ref{table:accuracy}, it is clear that the system performance improves with face masking because the eyes and mouth movement is completely obscured which avoids the noise due to movement. With the other experiment, we observed that the \acrshort{mae} was the lowest with natural daylight, thereby indicating that the light sources are adding artifacts to the video resulting in noise in the raw signal. Although the performance of the system is comparable on both the networks, the \acrshort{bptime} is similar but the \acrshort{srtime} varies for different devices and browsers. \acrshort{srtime} depends on data processing time plus the network data transmission time. The computational complexity, device configuration, and video streaming capability of the device camera affect the data processing time.  The frame rate of the camera in a browser depends on video resolution, device memory, and available bandwidth \cite{B62}. For example, a camera might capture at 60 \acrshort{fps} at 720p resolution but it might only capture 30 \acrshort{fps} at 1080p. %The number of applications running on a device also affects the frame rate. %The best back-end processing time and system response times were 46.4 \acrshort{s} and 91.26 \acrshort{s} and the worst times were 221.45 \acrshort{s} and 263.01 \acrshort{s}  respectively. Therefore, the back-end subsystem slows down as the number of users increase which is expected.

We configured the Gunicorn service with 5 worker processes as the official documentation of the service provider mentions that 4-12 worker processes can handle thousands of requests per second \cite{B63}. The recommendation is to use $(2 * \#num\_cores) + 1$ as the number of workers. Gunicorn relies on the operating system for load balancing. The Nginx server has efficient load balancing techniques that must be configured. The workload capacity of the framework can be enhanced by appropriate configurations, which is outside the scope of this work.

\begin{table}%[!b]
    \caption{Validating the framework accuracy.}
    \label{table:accuracy}
        \centering
        \small
        \begin{tabular}{|l|l|l|r|r|}
        \hline
            \textbf{Time} & \textbf{State} & \textbf{Vital} & \textbf{Mask (\acrshort{mae})} & \textbf{No Mask(\acrshort{mae})} \\
        \hline
            \multirow{4}{*}{Morning} & \multirow{4}{*}{Rest} & \acrshort{hr} & 9.4 \acrshort{bpm} & 7 \acrshort{bpm} \\ \cline{3-5}
                                                        &     & \acrshort{spo2} &  2.7\% & 3\% \\ \cline{3-5}
                                                        &     & \acrshort{sbp} &  8.1 mmHg & 10.3 mmHg \\ \cline{3-5}
                                                        &     & \acrshort{dbp} &  5.7 mmHg & 8.4 mmHg \\
        \hline
            \multirow{4}{*}{Morning} & \multirow{4}{*}{Post Exercise} & \acrshort{hr} & 6.3 \acrshort{bpm} & 9.7 \acrshort{bpm} \\ \cline{3-5}
                                                        &     & \acrshort{spo2} &  2.3\% & 2.3\% \\ \cline{3-5}
                                                        &     & \acrshort{sbp} &  9 mmHg & 7.3 mmHg  \\ \cline{3-5}
                                                        &     & \acrshort{dbp} &  3.6 mmHg & 3.3 mmHg \\
        \hline
            \multirow{4}{*}{Evening} & \multirow{4}{*}{Rest} & \acrshort{hr} & 5.25 \acrshort{bpm} & 8.25 \acrshort{bpm} \\ \cline{3-5}
                                                        &     & \acrshort{spo2} &  2.7\% & 2.7\% \\ \cline{3-5}
                                                        &     & \acrshort{sbp} &  11.75 mmHg & 15 mmHg \\ \cline{3-5}
                                                        &     & \acrshort{dbp} &  1.5 mmHg & 4 mmHg \\
        \hline
            \multirow{4}{*}{Evening} & \multirow{4}{*}{Post Exercise} & \acrshort{hr} & 6.3 \acrshort{bpm} & 6.6 \acrshort{bpm} \\ \cline{3-5}
                                                        &     & \acrshort{spo2} &  2.3\% & 2.3\% \\ \cline{3-5}
                                                        &     & \acrshort{sbp} &  9 mmHg & 10.6 mmHg \\ \cline{3-5}
                                                        &     & \acrshort{dbp} &  4.3 mmHg & 4 mmHg \\
        \hline
        \hline
            \multicolumn{2}{|l|}{\multirow{4}{*}{\textbf{Mean}}} & \textbf{\acrshort{hr}} & \textbf{6.8 \acrshort{bpm}} & \textbf{7.9 \acrshort{bpm}} \\ \cline{3-5}
                                          \multicolumn{2}{|l|}{} & \textbf{\acrshort{spo2}} &  \textbf{2.5\%} & \textbf{2.5\%} \\ \cline{3-5}
                                          \multicolumn{2}{|l|}{} & \textbf{\acrshort{sbp}} &  \textbf{9.4  mmHg} & \textbf{10.8  mmHg} \\ \cline{3-5}
                                           \multicolumn{2}{|l|}{}& \textbf{\acrshort{dbp}} &  \textbf{3.8  mmHg} & \textbf{4.9  mmHg} \\
        \hline

        \end{tabular}
    \end{table}

\begin{table}%[!b]
        \caption{Validating robustness of the framework with different light sources and camera resolutions.}
        \label{table:robustness}
            \centering
            \small
            \begin{tabular}{|l|r|r|r|r|r|r|}
            \hline
                & \multicolumn{3}{c|}{\textbf{Light(\acrshort{mae})}} & \multicolumn{3}{c|}{\textbf{Camera Resolution(\acrshort{mae})}} \\
            \hline
                 & \textbf{Daylight}  & \textbf{Warm Tone} & \textbf{Cool Tone} & \textbf{0.3MP}  & \textbf{2MP} & \textbf{8MP} \\
            \hline
                 \textbf{HR} & 2.1 \acrshort{bpm}  & 7 \acrshort{bpm} & 6.2 \acrshort{bpm} & 13.8 \acrshort{bpm}  & 6.1 \acrshort{bpm} & 8.2 \acrshort{bpm}   \\
            \hline
                 \textbf{SpO2} & 3\%  & 3\% & 3\% & 3\%  & 3\% & 3\%  \\
            \hline
                 \textbf{SBP} & 5.9 mmHg  & 9.8 mmHg & 5 mmHg & 2.1 mmHg  & 2 mmHg & 3.4 mmHg \\
            \hline
                \textbf{DBP} & 4.2 mmHg  & 6.1 mmHg & 5.8 mmHg &  6.1 mmHg & 1.4 mmHg & 3 mmHg  \\
            \hline

        \end{tabular}
\end{table}

\begin{table}%[!b]
    \caption{Validating the performance of the framework for different networks, devices and browsers.}
    \label{table:performance}
        \centering
        \small
        \begin{tabular}{|l|l|r|r|}

        \hline
             & & \textbf{Back-end processing time}  & \textbf{System response time}  \\
        \hline
             \multirow{2}{*}{Network} & \textbf{Wifi Network} & 47.9 \acrshort{s} & 84.5 \acrshort{s} \\ \cline{2-4}
             & \textbf{Mobile Network} & 50.1 \acrshort{s} & 86.0 \acrshort{s} \\
        \hline
        \hline
        \multirow{6}{*}{Devices} & \textbf{Acer laptop} & 50 \acrshort{s}  & 114 \acrshort{s}  \\ \cline{2-4}
        & \textbf{Alienware laptop} & 46.5 \acrshort{s}  & 104.5 \acrshort{s}  \\ \cline{2-4}
        
        & \textbf{Samsung phone} & 48.7 \acrshort{s}  & 107.5 \acrshort{s}  \\ \cline{2-4}
        
        & \textbf{Redmi phone} & 47.9 \acrshort{s}  & 116.7 \acrshort{s}  \\ \cline{2-4}
        
        &    \textbf{iPhone} & 48.3 \acrshort{s}  & 111.2 \acrshort{s} \\ \cline{2-4}
        
        &    \textbf{iPad} & 43 \acrshort{s} & 198 \acrshort{s} \\ \cline{2-4}
        \hline
        \hline
        \multirow{4}{*}{Browser} & \textbf{Google Chrome} & 48 \acrshort{s}  & 84.4 \acrshort{s}  \\ \cline{2-4}
        
        & \textbf{Mozilla Firefox} & 46.5 \acrshort{s}  & 104.55 \acrshort{s}  \\  \cline{2-4}
        & \textbf{Microsoft Edge} & 50.9 \acrshort{s}  & 93.47 \acrshort{s}  \\  \cline{2-4}
        
        & \textbf{Safari} & 46 \acrshort{s}  & 91.2 \acrshort{s}  \\
        \hline
    \end{tabular}
\end{table}

\section{Conclusion and Future Work}\label{sec:Conclusion}
With the global transition of business processes to online cloud technologies, the medical domain has seen a digital transformation in offering online services to patients. In this regard, the \acrshort{rppg} technology which facilitates measurement of vitals signs remotely from face videos can greatly benefit the online medical consultations. Existing research in \acrshort{rppg} \cite{B19,B26,B29} has shown promising results when used in controlled laboratory conditions. However, their performance degrades when movement or illumination changes affect the videos. Moreover, researchers have usually focussed on one or two vitals in validating their experiments \cite{B39,B40}. We propose a web-based, publicly accessible ubiquitous framework for estimating six vitals namely, \acrfull{hr}, \acrfull{hrv}, \acrfull{rr}, \acrfull{spo2}, \acrfull{bp}, and stress, which handles movement and illumination artifacts prevalent in real life. We validate the accuracy, robustness, usability and functionality of the \acrshort{rppg} models in estimating the vitals from face videos. %Further, we tested the application with different sources of light, and camera resolutions. The system performance was also validated on different networks at different times, using multiple browsers and devices. The application framework is compatible with all camera equipped devices having an internet connection.

As future work, ways to enhance the video quality need to be explored so that low resolution camera devices can be used over weak networks at remote locations. Better camera control can be used to optimize the frame rate for video capture. Videos can be live streamed using WebRTC technology to reduce processing delay instead of recording and uploading to the back end server. Further noise reduction due to facial movements can be explored in the future along with a study design to be executed at the hospital for a wider patient sample having varying vital sign measurements, which can help build a robust technology to deploy at a healthcare setting.  

%We developed and trained deep learning models for \acrshort{bp} estimation in this project. In deep learning models, uncertainty due to data and knowledge can be estimated and quantified using uncertainty models such as MC dropout, the Bootstrap model and the GMM \cite{B113}. This quantification makes the artificial intelligence systems more reliable and trustworthy. We need to perform this study with our deep learning models. We will also study the correlation between the prediction errors in estimating the different vital signs as all the values cumulatively project the cardiac activity.

\bibliographystyle{splncs04}
\bibliography{thesis}

\begin{thebibliography}{10}
\providecommand{\url}[1]{\texttt{#1}}
\providecommand{\urlprefix}{URL }
\providecommand{\doi}[1]{https://doi.org/#1}

\bibitem{B75}
Anura \url{https://www.anura.ai/} (accessed on February 16, 2022)

\bibitem{B77}
Binah.ai \url{https://www.binah.ai/} (accessed on February 16, 2022)

\bibitem{B52}
face-api.js, \url
  {https://justadudewhohacks.github.io/face-api.js/docs/index.html} (accessed
  on February 15, 2022)

\bibitem{B63}
Gunicorn 0.16.1 documentation, \url
  {https://docs.gunicorn.org/en/0.16.1/design.html} (accessed on February 15,
  2022)

\bibitem{B104}
Happitech \url{https://www.happitech.com/} (accessed on February 17, 2022)

\bibitem{B112}
Veyetals \url{https://veyetals.com/}(accessed on February 17, 2022)

\bibitem{B65}
Mediapipe face mesh  (2020), \url
  {https://google.github.io/mediapipe/solutions/face\_mesh.html} (accessed on
  February 15, 2022)

\bibitem{B47}
A.B., H.: The blood supply of various skin areas as estimated by the
  photoelectric plethysmograph. American Journal of Physiology  \textbf{124},
  328--340 (1938)

\bibitem{B68}
Castaneda~D., Esparza~A., G.M.S.C.N.H.: A review on wearable
  photoplethysmography sensors and their potential future applications in
  health care. Int J Biosens Bioelectron  (AUGUST 2018)

\bibitem{B16}
El-Hajj~C., K.P.: A review of machine learning techniques in
  photoplethysmography for the non-invasive cuff-less measurement of blood
  pressure. Biomedical Signal Processing and Control  \textbf{58 101870} (2020)

\bibitem{B39}
Fan~X., Ye~Q., Y.X.C.S.: Robust blood pressure estimation using an rgb camera.
  Journal of Ambient Intelligence and Humanized Computing  \textbf{11},
  4329--4336 (2018)

\bibitem{B24}
Guo~X., Li~Y., L.H.: Lime: Low-light image enhancement via illumination map
  estimation. IEEE Transactions on Image Processing  \textbf{26(2)},  982--993
  (February 2017)

\bibitem{B41}
Huang~P., Lin~C., C.M.L.T.W.B.: Image based contactless blood pressure
  assessment using pulse transit time. 2017 International Automatic Control
  Conference (CACS)  (2017), \url{doi:10.1109/cacs.2017.8284275 }(accessed on
  October 28, 2021)

\bibitem{B48}
Kamshilin~A., M.N.: Origin of photoplethysmographic waveform at green light.
  PNBS 2015

\bibitem{B93}
Kanva~A.K., Sharma~C.J., D.S.: Determination of spo 2 and heart-rate using
  smartphone camera. In Proceedings of The 2014 International Conference on
  Control, Instrumentation, Energy and Communication (CIEC) pp. 237--241
  (January 2014)

\bibitem{B36}
Kim~J., Cho~B., I.S.J.M.K.I.K.S.: Comparative study on artificial neural
  network with multiple regressions for continuous estimation of blood
  pressure. Int’l Conf. IEEE Eng. Med. and Bio pp. 6942--6945 (2005)

\bibitem{B19}
Kumar~M., Veeraraghavan~A., S.A.: Distanceppg: Robust non-contact vital signs
  monitoring using a camera. Biomedical optics express  \textbf{6(5)},
  1565--1588 (2015)

\bibitem{B62}
L., J.: What are webcam frame rates (JULY 2020), \url
  {https://www.lifewire.com/webcam-frame-rates-2640479} (accessed on February
  15, 2022)

\bibitem{B26}
Li~X., Chen~J., Z.G.P.M.: Remote heart rate measurement from face videos under
  realistic situations. In Proceedings of the IEEE conference on computer
  vision and pattern recognition pp. 4264--4271 (2014)

\bibitem{B29}
Nemcovaa~A., Jordanovaa~I., V.M.S.R.M.L.S.L.V.M.: Monitoring of heart rate,
  blood oxygen saturation, and blood pressure using a smartphone. Biomedical
  Signal Processing and Control  \textbf{59} (2020)

\bibitem{B102}
Park~C., L.B.: Real-time estimation of respiratory rate from a
  photoplethysmogram using an adaptive lattice notch filter. BioMedical
  Engineering Online  (December 2014)

\bibitem{B50}
Poh~M., McDuff~D., P.R.: Non-contact, automated cardiac pulse measurements
  using video imaging and blind source separation. Optics Express  \textbf{18},
   10762--10774 (2010)

\bibitem{B59}
Qiao~D., Zulkernin, F.M.R.R.R.J.N.: Measuring heart rate and heart rate
  variability with smartphone camera. 22nd IEEE International Conference on
  Mobile Data Management (MDM)  (2021)

\bibitem{B49}
Qiao~D., Ayesha~A., Z.F.M.R.J.N.: Revise: Remote vital signs measurement using
  smartphone video camera  (2022), \url{ https://arxiv.org/abs/2206.08748 }

\bibitem{B46}
Rahman~H., Ahmed~M., B.S.F.P.: Real time heart rate monitoring from facial rgb
  color video using webcam. 9th Annual Workshop of the Swedish Artificial
  Intelligence Society (SAIS)  (2016)

\bibitem{B4}
Shaffer~F., G.J.P.: An overview of heart rate variability metrics and norms
  (September 2017)

\bibitem{B37}
Shimazaki~S., Bhuiyan~S., K.H.O.K.: Features extraction for cuffless blood
  pressure estimation by autoencoder from photoplethysmography. 40th Annual
  International Conference of the IEEE Engineering in Medicine and Biology
  Society  (2018)

\bibitem{B8}
Sun~Y., T.N.: Photoplethysmography revisited: from contact to noncontact, from
  point to imaging  \textbf{63(3)},  463--477 (March 2016)

\bibitem{B40}
Viejo~C., Fuentes~S., T.D.D.F.: Non-contact heart rate and blood pressure
  estimations from video analysis and machine learning modelling applied to
  food sensory responses: A case study for chocolate. Sensors  \textbf{18(6)}
  (2018), \url{https://www.mdpi.com/1424-8220/18/6/1802}(accessed on September
  25, 2021)

\bibitem{B22}
Viola~P., J.M.: Rapid object detection using a boosted cascade of simple
  features. Proceedings of the 2001 IEEE Computer Society Conference on
  Computer Vision and Pattern Recognition  \textbf{1},  511--518 (2001)

\bibitem{B1}
Wang~C., Pun~T., C.G.: A comparative survey of methods for remote heart rate
  detection from frontal face videos  (2018)

\bibitem{B80}
Wang~Z., Yang~X., C.K.: Accurate face alignment and adaptive patch selection
  for heart rate estimation from videos under realistic scenarios. PloS one
  \textbf{13(5)} (2018)

\end{thebibliography}
\end{document}